%% file: main.tex
\documentclass[conference]{IEEEtran}

\usepackage{cite}
\usepackage{amsmath,amssymb,amsfonts}
\usepackage{algorithmic}
\usepackage{graphicx}
\usepackage{textcomp}
\usepackage{xcolor}
\usepackage{mathtools}

\usepackage{dsfont}
\usepackage{caption}
\usepackage{subcaption}
\usepackage{tabularx,booktabs}
\usepackage{slashbox}
\usepackage{booktabs,caption}
\usepackage[flushleft]{threeparttable}
\usepackage{multirow}

\usepackage{amsmath}
\usepackage{color}

\usepackage{tabularx}

\newcolumntype{C}[1]{>{\centering\arraybackslash}p{#1}}

\begin{document}
\title{Bitcoin Volatility Forecasting with a Glimpse into Buy and Sell Orders}

\author{\IEEEauthorblockN{Tian Guo}
\IEEEauthorblockA{ETH, Z\"urich, Switzerland\\
Email: tian.guo@gess.ethz.ch}
\and
\IEEEauthorblockN{Albert Bifet}
\IEEEauthorblockA{T\'el\'ecom ParisTech, France\\
Email: albert.bifet@telecom-paristech.fr}
\and
\IEEEauthorblockN{Nino Antulov-Fantulin}
\IEEEauthorblockA{ETH, Z\"urich, Switzerland\\
Email: nino.antulov@gess.ethz.ch}
}


\maketitle

\begin{abstract}
Bitcoin is one of the most prominent decentralized digital cryptocurrencies, currently having the largest market capitalization among cryptocurrencies. 
Ability to understand which factors drive the fluctuations of the Bitcoin price and to what extent they are predictable is interesting both from theoretical and practical perspective. 
In this paper, we study the problem of the Bitcoin short-term volatility forecasting by exploiting volatility history and order book data.
Order book, consisting of buy and sell orders over time, reflects the intention of the market and is closely related to the evolution of volatility. 
We propose temporal mixture models capable of adaptively exploiting both volatility history and order book features for short-term volatility forecasting.  
By leveraging rolling and incremental learning and evaluation procedures, we demonstrate the prediction performance of our model as well as studying the robustness, in comparison to a variety of statistical and machine learning baselines. 
Meanwhile, our temporal mixture model enables to decipher time-varying effect of order book features on the volatility.
It demonstrates the prospect of our temporal mixture model as an interpretable forecasting framework over heterogeneous Bitcoin data.
\end{abstract}

%
%




\input{introduction}
\input{related_work}

\input{preliminary}
\input{model}

\input{train}
\input{experiments}

\input{conclusion}

\section*{Acknowledgement}
The work of T.G. and N.A.-F. has been funded by the EU Horizon 2020 SoBigData project under grant agreement No. 654024.

\bibliographystyle{IEEEtran}
\bibliography{sample-bibliography} 

\end{document}

%% file: introduction.tex
\section{Introduction}\label{sec:intro}
Bitcoin (BTC) is a digital currency system which functions without central governing authority \cite{Nakamoto2008}.
It originated from a decentralized peer-to-peer payment platform through the Internet.
When new transactions are announced on this network, they have to be verified by network nodes and recorded in a public distributed ledger called the blockchain. 
Bitcoins are created as a reward in the verification competition in which users offer their computing power to verify and record transactions into the blockchain. 

Bitcoins can also be exchanged for other currencies, products, and services.
The exchange of the Bitcoins with other currencies is done in the exchange office, where "buy" or "sell" orders are stored on the
order book \cite{Nakamoto2008,Ns2006}. 
"Buy" or "bid" orders represent an intention to buy a certain amount of Bitcoins at some maximum price while "sell" or "ask" orders represent an intention to sell a certain amount of Bitcoins at some minimum price. The exchange is done by matching orders by price from the order book into a trade transaction between buyers and sellers. 

Due to Bitcoin's growing popular appeal, numerous studies have been conducted recently to identify statistical or economical properties and characterizations of Bitcoin.   
For instance, these research focus on statistical properties \cite{Chu2015}, 
bubbles in Bitcoin \cite{cheah2015speculative, Garcia2014}, insight into the market crash~\cite{BouchaudBTC}, 
the relationship between Bitcoin and web information, such as Google Trends and Wikipedia \cite{kristoufek2013bitcoin}, and wavelet analysis of Bitcoin~\cite{kristoufek2015main}.
However, there are few papers analyzing the Bitcoin processes in terms of prediction performance from order book data.


To this end, in this paper, we focus on studying the predictive performance of Bitcoin price short-term volatility using both volatility history and order book data.
Though Bitcoin is the largest of its kind in terms of total market capitalization value, it still suffers from a volatile price.
Volatility as a measure of price fluctuations \cite{Andersen2003, Hansen2005} has a significant impact on trade strategies, investment decisions \cite{Fleming2003} as well as systemic risk \cite{Pikorec2014}.
Meanwhile, order book data carrying fined-grained information about price movement and market intentions is proven to be closely related to volatility \cite{Ns2006}
and influences Bitcoin market with variation over time \cite{BouchaudBTC}.
Therefore, it is of great interest to data mining and machine learning community to be able to develop predictive models for Bitcoin volatility as well as characterizing the time-varying impact of order book on the volatility evolution.
Note that this paper has no intention to produce another financial models of volatility \cite{Andersen2003, Hansen2005} or the limit order book dynamics itself \cite{OB_rev_Porter2013}.
\begin{figure}[!htbp]
\centering
\includegraphics[width=0.42\textwidth]{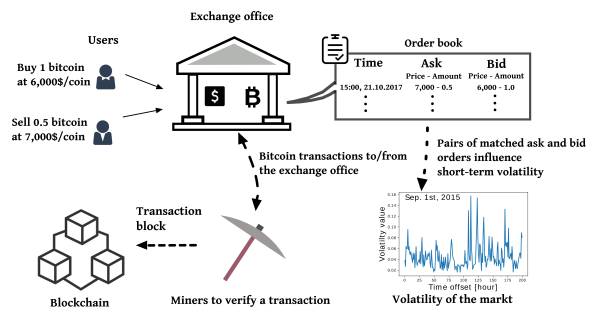}
\caption{Intuitive illustration of Bitcoin orders in an exchange office. Users announce buy or sell orders to the exchange office, which are recorded in the order book. Orders reflect the intention of the market and influence the volatility. Bitcoin transactions to or from the exchange office are broadcast to miners for verification. 
}
\label{fig:btc}
\end{figure}

Specifically, the main contributions are: 
(1) we formulate the Bitcoin volatility forecasting problem as learning predictive models over volatility history and features extracted from order book; 
(2) we propose probabilistic temporal mixture models to capture autoregressive dependence in volatility history as well as the dynamical effect of order book features on the volatility evolution; 
(3) with the rolling and incremental evaluation methodologies, we not only demonstrate the superior prediction performance of our temporal mixture model in comparison to a variety of statistical and machine learning baselines but also study the robustness of each approach w.r.t. the look-back horizon of training data. 
Our experiments provide a comprehensive evaluation of prediction models for Bitcoin volatility;
(4) by visualizing the gate value and important features over time obtained in the temporal mixture model, we detect regimes when order book presents high impact on volatility and provide intuitive interpretation.
Note that by adding component models specific to additional data sources, for instance, Blockchain data \cite{jang2018empirical}, social media \cite{kim2016predicting}, different exchange offices, etc, our temporal mixture model is able to serve as a unified framework for forecasting and studying the effect of different data sources on volatility. However, they are out of the scope of current work and are left for future work. 


The remainder of the paper is structured as follows.
In Section \ref{sec:related}, we give an overview of the related work. Section \ref{ref:pre} gives the detailed description of Bitcoin data and problem formulation. 
Section \ref{sec:model} and \ref{sec:train} explain the proposed temporal mixture model and associated learning and evaluation methodology.
Finally, in section \ref{sec:exp}, we report the experimental results, followed by the conclusion \ref{sec:conclusion}. 


%% file: related_work.tex
\section{Related work} \label{sec:related}
Different studies have tried to explain various aspects of the Bitcoin such as its price formation, volatility, systems dynamics and economic value. 
From the economic perspective, main studies \cite{Chu2015, cheah2015speculative} were focused on understanding the fundamental and speculative value of Bitcoin. \cite{Garcia2014} exploited autoregression techniques to identify positive feedback loops leading to price bubbles. 
In data mining and machine learning models areas, \cite{amjad2017trading, alessandretti2018machine} used the historical price time series for price prediction and trading strategies.
\cite{li2018sentiment, kim2016predicting} utilized social information like the sentiment, comments, and replies on forums to forecast price fluctuations. \cite{jang2018empirical} explored the predictive ability of Blockchain information for Bitcoin price.
As for volatility prediction, \cite{Katsiampa2017} evaluated the performance of GARCH models on Bitcoin.
However, volatility forecasting using order book information of Bitcoin is still under-researched. In this paper, we develop predictive models consuming volatility history and order book information.

Meanwhile, numerous studies have been done for forecasting stock price, return and fluctuation by using different data sources, e.g. price history or social media data. 
\cite{zhang2017stock, rather2015recurrent, jiao2017predicting, campbell2007predicting} developed machine learning and recurrent neural network based approaches using price historical time series.
\cite{si2013exploiting, hu2017listening, ding2015deep} extracted sentiment and event features from Twitter and news for stock market prediction. 
Some recent work attempted to exploit heterogeneous historical price and social media data via feature concatenation \cite{nguyen2015topic} or joint feature learning \cite{xustock} for stock prediction.
When such feature fusion methods are applied to the problem of forecasting volatility with the assistance of order book data in this paper, they overlook the time-varying environment of the market \cite{BouchaudBTC, kristoufek2015main, kristoufek2013bitcoin} as well as weakening the interpretability of order book features \cite{oyamada2017relational}.

Instead, in this paper, we propose interpretable temporal mixture models, which are aimed at improving prediction performance as well as exhibiting the time-varying interaction between volatility and Bitcoin orders. Compared to recurrent neural network based methods, our model has well-defined data flows and architecture to capture time-varying effect in data.


Another line of related research is about mixture models in various classification and regression problems. 
Proposed by \cite{jordan1994hierarchical}, mixture models or the mixture of experts is a learning paradigm that divides one task into a subset of distinct tasks (i.e., expert), and then utilizes a gating function to weight the output of individual tasks \cite{yuksel2012twenty}. 
\cite{wei2007dynamic} presented dynamic mixture models for online pattern discovery in time series.
\cite{oyamada2017relational} developed mixture models to analyze behavioral data of customers for demographics prediction.
In this paper, we enhance the mixture model by developing temporal gate function and hinge regularizations. 
To the best of our knowledge, this is the first work to exploit mixture models for predicting Bitcoin volatility.

%% file: preliminary.tex
\section{Volatility history and order book}\label{ref:pre}
In this paper, we collect the realized volatility history of Bitcoin, which refers to the standard deviation of returns within a short time interval \cite{Hansen2005}. 
The return is defined as the relative change in consecutive prices of BTC. 
This dataset contains time series of hourly volatility spanning more than one year from the OKCoin, which is an exchange office platform providing trading services between fiat currencies (USD, EUR, CNY) and cryptocurrencies.
During the period of the collected data, trading volume of BTC at the OKCoin exchange office was approximately $40\%$ of the total traded BTC volume,  which implies that our data source (OkCoin) can be used as a good proxy.

\begin{figure}
\centering
\includegraphics[width=0.44\textwidth]{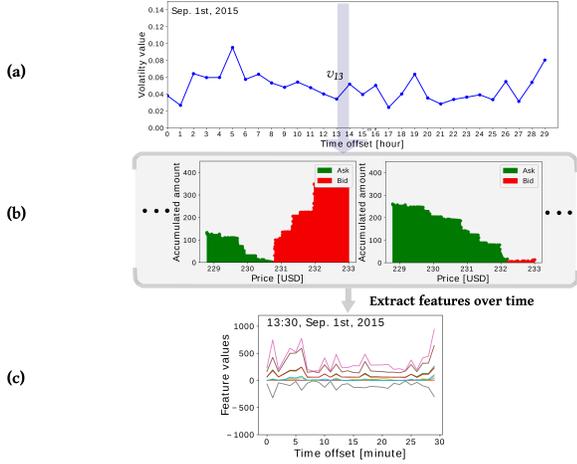}
\caption{Volatility observations with order book features.
Top panel: volatility series 
on time offset in hours w.r.t. 00:00AM, September 1st, 2015.
Middle panel: order book snapshots on minutes within hour $13$.
Bottom panel: order book snapshots are transformed to series of features.
}
\label{fig:feature}
\end{figure}

In addition, we have the order book data from OKCoin over the same period of volatility.
It was collected through the exchange API with the granularity of one minute, with negligible missing values due to the API downtime or communication errors.
Each order contains the \textit{amount} of Bitcoins customers intend to buy or sell at corresponding \textit{price}.
For instance, the middle panel of Fig. \ref{fig:feature} shows two snapshots of buy and sell orders. 
The green and red areas respectively show the accumulated amount on the ask and bid sides w.r.t. the prices. This figure is commonly used to interpret the market intention and potential movement. 





The volatility series outlines the long-term fluctuation of Bitcoin price over time, while order book data provides fine-grained selling and buying information characterizing the instantaneous local behavior of the market.
Therefore, for forecasting the volatility it is highly desirable to develop a systematical way to model the complementary dependencies in volatility history and orders. 
Intuitively, our idea is to first transform order book data into features over time and then to develop probabilistic models over volatility and order book features.

Concretely, from each snapshot of order book data, we extract the order book features \cite{OB_rev_Porter2013} such as: volume, depth, spread and slope for bid and ask sides. 
\begin{itemize}
    \item \textit{spread} is the difference between the highest price that a buyer is willing to pay for a BTC (bid) and the lowest price that a seller is willing to accept (ask).  
    \item \textit{Ask/Bid Depth} is the number of orders on the bid or ask side. 
    \item \textit{Depth difference} is the difference between ask and bid depth. 
    \item \textit{Ask/Bid Volume} is the number of BTCs on the bid or ask side.
    \item \textit{Volume difference} is the difference between ask and bid volume. 
    \item \textit{Weighted spread} is the difference between cumulative price over $10\%$ of bid depth and the cumulative price over $10\%$ of ask depth. 
    \item \textit{Ask/bid Slope} is estimated as the volume until $\delta$ price offset from the current traded price, where $\delta$ is estimated by the ask (or bid) price at the order that has at least $10\%$ of orders with the higher ask (or bid) price. 
\end{itemize}


For instance, Fig. \ref{fig:feature} illustrates how volatility observations and the associated order book features are organized. The shaded area in the top panel demonstrates time period corresponding to volatility $v_{13}$. The bottom panel shows that the order book data at each snapshot within the time range of the shaded area are transformed into feature series, which will be fed into prediction models.

We formulate the problem to resolve in this paper as follows. Given a series of volatility observations $\{ v_0, \ldots, v_H \}$, the $h$-th observation is denoted by $v_h \in \mathbb{R}^{+}$. 
The features of the order book at each snapshot are denoted by a vector $\mathbf{x}_{m} \in \mathbb{R}^{n}$, where $n$ is the dimension of the feature vector. 
We define an index mapping function $i( \cdot )$ to map the time index $h$ of a volatility observation to the last time index of order book snapshot before $h$. 
For instance, in our present dataset, volatility and order book snapshot are respectively hourly and minutely data. Thus, for $h=1$, $i(h)=60$. 
Now order book features associated with a volatility observation $v_{h}$ is denoted by a matrix $\mathbf{X}_{ [i(h) \, , -l_b ]} = (\mathbf{x}_{i(h)}, \ldots, \mathbf{x}_{i(h) - l_b -1} )  \in R^{n \times l_b}$, where $l_b$ is the time horizon. Likewise, a set of historical volatility observations w.r.t. $v_{h}$ is denoted by $\mathbf{v}_{[h \, , -l_v]} = ( v_{h-1}, \ldots, v_{h-l_v} ) \in \mathbb{R}^{l_v}$. Given historical volatility observations $\mathbf{v}_{[h \, , -l_v]}$ and order book features $\mathbf{X}_{ [i(h) \, , -l_b ]}$, we aim to predict the $D$-step ahead volatility $\hat{v}_{h+D}$. In addition, the proposed model should be interpretable, in the sense that it enables to observe how such two types of data interact to drive the evolution of volatility.

%

%% file: model.tex
\section{Models}\label{sec:model}
Time series model is the natural choice for volatility history observations.
It can be modeled via the classical autoregressive and integrated moving average model (ARIMA).
As for the order book features, they can be incorporated as exogenous variables into ARIMA by adding regression terms on the features \cite{hyndman2014forecasting}. This gives rise to ARIMAX model, which assumes a static relation between order book features and the volatility via the regression terms.
However, the following proposed temporal mixture model is aimed at adaptively exploiting volatility history and order book features for forecasting as well as characterizing the impact of order book over time. 
\subsection{Temporal mixture model}

A mixture model is a weighted sum of component models \cite{yuksel2012twenty}.
Individual component models specialize on the different part of the data. 
The weights dependent on input data enable the model to adapt to non-stationary data \cite{yuksel2012twenty, jordan1994hierarchical}.

Our temporal mixture model starts with building a joint probabilistic density function of volatility observations conditional on order book features as follows:
\begin{equation}\label{formula:joint}
\begin{split}
& p( v_1, \ldots, v_H \, | \, \{\mathbf{x}_m\} ; \Theta ) = \\
& \sum_{z_1} \cdots \sum_{z_H} p(v_1, \ldots, v_H \,, z_1, \ldots, z_H | \, \{\mathbf{x}_m\} ; \Theta) = \\
& \prod_{h} \Big[ p( v_h \, , \, z_h = 0 \, | \, \mathbf{v}_{[h, -l_v]} \, , \, \mathbf{X}_{ [i(h), -l_b ]}) \\
& + p( v_h \, , \, z_h = 1 \, | \, \mathbf{v}_{[h, -l_v]} \, , \, \mathbf{X}_{ [i(h), -l_b ]}) \Big].
\end{split}
\end{equation}
In Eq. \ref{formula:joint}, we introduce a binary random latent variable $z_h \in \{0, 1\}$ for each observation, which corresponds to two cases of the density of $v_h$ conditional on historical volatility and order book feature (i.e. $\mathbf{v}_{[h, -l_v]}$ and $\mathbf{X}_{ [i(h), -l_b ]}$).
$z_h = 0$ corresponds to the case when volatility is driven by the historical data, while $z_h = 1$ stands for the dependence on order book features.
Then, the joint density function of $( v_1, \ldots, v_H, z_1, \ldots, z_H )$ is decomposed into the product of the density of individual observations.
$\Theta$ is the set of parameters, which is shown below.

\begin{figure}
\centering
\includegraphics[width=0.3\textwidth]{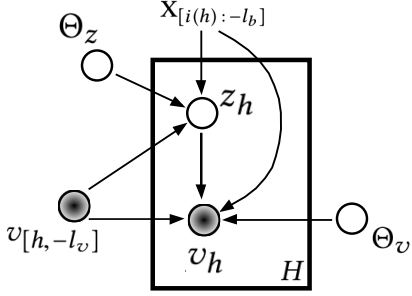}
\caption{Graphical representation of the temporal mixture model.}
\label{fig:graph}
\end{figure}

Concretely, by defining the conditional probability of latent variable $z_h$ as $g_h$, Eq. \ref{formula:joint} is rewritten as:
\begin{equation}\label{formula:comp}
\begin{split}
p(v_1, \ldots, v_H \, & | \, \{\mathbf{x}_m\} ; \Theta)  = \prod_{h} \Big[ p( v_h \, | \, \mathbf{v}_{[h \, ,-l_v]}, z_h = 0 )\cdot g_h \\
& + p( v_h \, | \, \mathbf{X}_{[i(h), -l_b ]}, z_h = 1 ) \cdot (1-g_h) \Big],
\end{split}
\end{equation}
where $g_h \coloneqq P(z_h = 0 \, | \, \mathbf{v}_{[h \, ,  -l_v]}, \, \mathbf{X}_{[i(h), -l_b ]} )$.
The mixture components $p( v_h \, | \, \mathbf{v}_{[h \, ,-l_v]}, z_h = 0 )$ and $p( v_h \, | \, \mathbf{X}_{[i(h), -l_b ]}, z_h = 1 )$ define individual density functions of $v_h$ respectively conditional on volatility history and order book features. 

Through the gate function $g_h$, latent variable $z_h$ governs the generation of $v_h$ based on $\mathbf{v}_{[h \, , -l_v]}$ and $\mathbf{X}_{[i(h), -l_b ]}$.
In particular, $g_h$ represents the weight for the volatility history component, 
where $v_h$ is driven solely by its history i.e. $p( v_h \, | \, \mathbf{v}_{[h \, ,-l_v]}, z_h = 0 )$, 
while $1-g_h$ is the weight for the order book component  $p( v_h \, | \, \mathbf{X}_{[i(h), -l_b ]}, z_h = 1 )$.
As a result, data dependent $g_h$ adaptively adjusts the contribution from history and order book to $v_h$.
Note that if there are additional data sources fed into the temporal mixture model, it can simply accommodate them by expanding the domain of $z_h$ and adding corresponding component models.  

In the inference phase, two component models derive mean value conditioned on their respective input.
The weighted combination of their means by $g_h$ is taken as the prediction of the mixture model.
Moreover, this temporal mixture model is interpretable in the sense that by observing the gate values, we can understand when and to what extent order book contributes to the evolution of volatility. We demonstrate this in the experiment section.

Considering the characteristics of volatility data, we describe two realizations of the temporal mixture model  based on Gaussian and log-normal distributions as follows. 

\subsection{Gaussian temporal mixture model}
In this part, we choose the Gaussian distribution to model the conditional density of $v_h$ under different states of latent variable $z_h$. 
Specifically, they are represented as:
\begin{equation}
\begin{split}
v_h \, | \, \{\mathbf{v}_{[h \, , -l_v]} \, , z_h = 0 \} \sim \mathcal{N}( v_h | \, \mu_{h,0} \, , \, \sigma^{2}_{h,0}  ) \, \\ 
v_h \, | \, \{\mathbf{X}_{[i(h) \, , -l_b]} \, , z_h = 1 \} \sim \mathcal{N}( v_h | \, \mu_{h, 1}, \, \sigma^{2}_{h, 1} ), \, 
\end{split}
\end{equation}
where $\mu_{h, \cdot}$ and $\sigma^{2}_{h, \cdot}$ are the mean and variance of individual Gaussian distributions.
\begin{equation}\label{formula:mean}
\begin{split}
\mu_{h,0} &= \sum \phi_{j} v_{h-j} \\
\mu_{h,1} &= U^\top \mathbf{X}_{[i(h) \, : \, -l_b]} V,
\end{split}
\end{equation}
where $\phi_{i} \in \mathbb{R}$, $ U \in \mathbb{R}^n$ and $ V \in \mathbb{R}^{l_b}$ are the parameters to learn. 

In Eq. \ref{formula:mean}, we use an autoregressive model to capture the dependence of $v_h$ on historical volatility. Order book features are organized as a matrix with temporal and feature dimensions. Therefore we make use of bilinear regression, where parameters $U$ and $V$ respectively capture the temporal and feature dependence.
As a result, feature importance can be interpreted with ease, which is illustrated in the experiment section. The variance term $\sigma^{2}_{h, \cdot}$ in each component is obtained by performing linear regression on the input of that component or set to constant. 

Then, the gate function $g_h$ is defined by the softmax function
\begin{equation}\label{formula:softmax}
g_h \coloneqq \frac{\exp(\, \sum \theta_{j} v_{h-j} \, )}{ \exp( \, \sum \theta_{j} v_{h-j} \, ) + 
\exp( \, A^\top \mathbf{X}_{[i(h) \, , \, -l_b]} B \, )},
\end{equation}
where $\theta_{i} \in \mathbb{R}$, $ A \in \mathbb{R}^n $ and $ B \in \mathbb{R}^{l_b} $ are the parameters to learn. 
Likewise, we utilize autoregression and bilinear regression, thereby facilitating the understanding of the feature importance in determining the contribution of volatility history and order book features.

During the inference, the conditional mean of the mixture distribution is taken as the predicted value $\hat{v}_{h}$ :
\begin{equation}\label{eq:pred}
\begin{split}
\hat{v}_{h} &= \mathbb{E}\left(v_{h} | \mathbf{v}_{[h \, , -l_v]} \, , \, \mathbf{X}_{[i(h) \, , -l_b]}  \right) \\ 
&= g_h \cdot \mu_{h,0} + (1-g_h) \cdot \mu_{h,1}. 
\end{split}
\end{equation}

We define $\Theta \coloneqq \{ \phi_{i}, U, V, \theta_{i}, A, B \}$ as the entire set of parameters in the mixture model. We present the loss functions for learning 
$\Theta$ and in the next section we describe the detailed learning algorithm.

In learning the parameters in the Gaussian temporal mixture model, the loss function to minimize is defined as:
\begin{equation}\label{formula:obj}
\begin{split}
\mathcal{O}(\Theta) \coloneqq & - \mathcal{L}_g 
 + \lambda \left\Vert  \Theta \right\Vert^2_2 + \\
& \underbrace{ \alpha \sum_{h} \left[ \max(0 \, , \, \delta - \mu_{h,0} ) + \max(0 \, , \, \delta - \mu_{h,1})  \right] }_{\text{Non-negative mean regularization}} ,  
\end{split}
\end{equation}
where 

$\mathcal{L}_g = \sum_h \log \left[ g_h \mathcal{N}(v_h \, | \, \mu_{h,0}, \sigma^{2}_{h,0} ) + (1 - g_h)\mathcal{N}(v_h \, | \, \mu_{h,1}, \sigma^{2}_{h,1} ) \right]$ is the log likelihood of volatility observations.
In addition to the L2 regularization over $\Theta$ for preventing over-fitting, we introduce two hinge terms to regularize the predictive mean of each component model, i.e. $\mu_{h,0}$ and $\mu_{h,1}$. 
This is because the value of volatility lies in the non-negative domain of real values and we impose hinge loss on the mean of each component model to penalize negative values. 
The parameter $\delta$ is the margin parameter, which is empirically set to zero in experiments. 

\subsection{Log-normal temporal mixture model}
Instead of enforcing the non-negative mean of a component model by regularization,
in this part, we present the temporal mixture model using log-normal distribution, which naturally fits non-negative values \cite{Andersen2003}. 

Specifically, for a random non-negative variable of log-normal distribution, the logarithm of this variable is normally distributed.
Thus, by assuming $v_h$ is log-normally distributed, we represent component models of the temporal mixture model as:
\begin{equation}
\begin{split}
\log(v_h) \, | \, \{ \mathbf{v}_{[h \, , -l_v]} \, , z_h = 0 \} \sim \mathcal{N}( \log(v_h) | \, \mu_{h,0} \, , \, \sigma^{2}_{h,0}  ) \, 
\\
\log(v_h) \, | \, \{ \mathbf{X}_{[i(h) \, , -l_b]} \, , z_h = 1 \} \sim \mathcal{N}( \log(v_h) | \, \mu_{h, 1}, \, \sigma^{2}_{h, 1} ). \, 
\end{split}
\end{equation}
The conditional mean of $v_h$ in such component models becomes $ \mathbb{E}(v_h \, | \cdot) = \exp( \mu + 0.5\sigma) $.
Regarding the function $g_h$, we use the same form as in Eq. \ref{formula:softmax}.
In the loss function, we can safely get rid of the non-negative regularization, due to the non-negative nature of $\mathbb{E}(v_h \, | \cdot)$ and obtain the loss function as:
\begin{equation}\label{formula:obj-log}
\begin{split}
\mathcal{O}(\Theta) \coloneqq & -  \mathcal{L}_{log} + \lambda \left\Vert  \Theta \right\Vert^2 ,
\end{split}
\end{equation}
where $\mathcal{L}_{log}$ is the log likelihood of volatility observations defined by the log-normal distribution, i.e.
$\mathcal{L}_{log} = \sum_{h} \log \left[ g_h \cdot p(v_h \, | \, \mu_{h,0}, \sigma^{2}_{h,0} ) + (1-g_h) \cdot p(v_h \, | \, \mu_{h,1}, \sigma^{2}_{h,1} ) \right]$.
$ p(v_h \, | \, \cdot )$ is the density function of log-normal. Due to the limitation of pages, we skip the details of log-normal distribution.



%% file: train.tex
\section{Model Learning and Evaluation}\label{sec:train}
In this section, we describe the learning algorithm for the temporal mixture models as well as the evaluation scheme. 

\subsection{Learning methods}
Our mixture model involves both latent states and coupled parameters and thus we 
iteratively minimize the objective function defined in Eq.\ref{formula:obj} and Eq.\ref{formula:obj-log} \cite{yuksel2012twenty}. We use the Gaussian temporal mixture model to illustrate the algorithm, while the same methodology is applied to the log-normal temporal mixture model as well. 

Specifically, the learning algorithm consists of two main steps. First, fix all component model parameters and {update the parameters of the gate function} $g_h$, by using a gradient descent method \cite{ murphy2012machine}. Due to the page limitation, We present the gradients of certain parameters w.r.t. the objective function below. The rest can be derived analogously.
\begin{equation}
\begin{split}
\frac{\partial \mathcal{O}}{\partial \theta_i} &= -\sum_{h} \Big[ \frac{1}{a(v_h \, | \, \Theta)} g_h v_{h-i} (1-g_h) \mathcal{N}(v_h \, | \, \mathbf{v}_{[h, -l_v]} \, ) \\
& - g_h v_{h-i} (1-g_h) \mathcal{N}(v_h \, | \, \mathbf{X}_{[i(h), -l_b]}) \Big]
+ 2\lambda \theta_i,
\end{split}  
\end{equation}
where $a(v_h \, | \, \Theta)$ denotes $ g_h \mathcal{N}(v_h \, | \, \mu_{h,0}, \sigma^{2}_{h,0} ) + (1-g_h)\mathcal{N}(v_h \, | \, \mu_{h,1}, \sigma^{2}_{h,1} )$.
For simplicity, in the following formulas, we ignore the value of $z_h$ in the conditions.

For the coupled parameters of bilinear regression, it can be broken into two convex tasks, where we individually learn parameters as follows \cite{shi2014sparse}:
\begin{equation}
\begin{split}
\frac{\partial \mathcal{O}}{\partial A} &=  -\sum_{h}\frac{\mathbf{X}_{[i(h) \, , \, -l_b]} B}{a(v_h \, | \, \Theta)} \Big[ \frac{(g_h-1)g_h }{\exp( \, \sum \theta_{i} v_{h-i} \, )} \mathcal{N}(v_h \, | \, \mathbf{v}_{[h, -l_v]} )  \\ 
& -\frac{(g_h-1)g_h }{\exp( \, \sum \theta_{i} v_{h-i} \, )} \mathcal{N}(v_h \, | \, \mathbf{X}_{[i(h), -l_b]} )
\Big] + 2\lambda A.
\end{split}  
\end{equation}


Second, fix the gate function and {update the parameters in component models}:  
$P( v_h \, | \, v_{[h \, , -l_v]} \, , z_h = 0 )$ and \\$P( v_h \, | \, \mathbf{X}_{[i(h) \, , -l_b]} \, , z_h = 1 )$. 
\begin{equation}
\begin{split}
\frac{\partial \mathcal{O}}{\partial \phi_i} = &-\sum_{h} \frac{2 g_h \cdot \mathcal{N}(v_h \, | \, \mathbf{v}_{[h, -l_v]} ) }{a(v_h \, | \, \Theta) \sigma^{2}_{h,0}} \left( \sum_{j} \phi_j v_{h-j} - v_h  \right) v_{h-i} \\ & + 2\lambda \phi_i + \alpha \sum_{h} \mathds{1}_{>0}\{\max(0 \, , \, \delta - \mu_{h,0})\}  (-v_{h-i}).
\end{split}  
\end{equation}
\begin{equation}
\begin{split}
& \frac{\partial \mathcal{O}}{\partial V} =  + 2\lambda V + \alpha \sum_{h} \mathds{1}_{>0}\{\max(0 \, , \, \delta - \mu_{h,1})\} (-\mathbf{X}_{[i(h), -l_b]}^\top U)\\
& -\sum_{h} \frac{2(1-g_h) \cdot \mathcal{N}(v_h \, | \, \mathbf{X}_{[i(h), -l_b]})}{a(v_h \, | \, \Theta)  \sigma^{2}_{h,1}} \left( U^\top \mathbf{X}_{[i(h), -l_b]}V - v_h \right) \\ 
& \cdot \mathbf{X}_{[i(h), -l_b]}^\top U.
\end{split}  
\end{equation}

\subsection{Evaluation procedure}
The standard procedure of learning and evaluating time series models is to split the entire time series at a certain time step. Then the front part is taken as training and validation data, while the rest is used as testing data \cite{hooi2017b, laptev2017time}.
Bitcoin data is non-stationary in the sense that old data could differ from the recent one in terms of statistical characteristics \cite{amjad2017trading, Katsiampa2017}.
As a result, the aforementioned learning process using all the data preceding to the testing period has to compromise the non-stationarity in data and leads to degraded prediction performance.
Therefore, we adopt a rolling strategy to learn and evaluate models \cite{amjad2017trading, jang2018empirical}. It enables to study the performance of models on different time periods of the data. 
\begin{figure}[!htbp]
\centering
\includegraphics[width=0.4\textwidth]{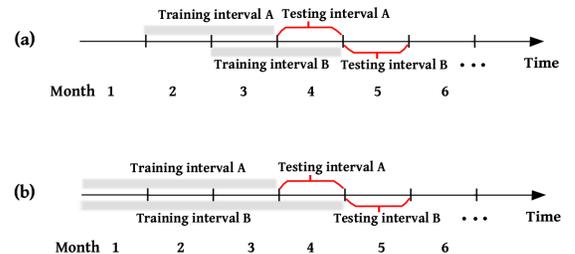}
\caption{Top panel: rolling procedure. Bottom panel: incremental procedure.}
\label{fig:eval}
\end{figure}

The process of \textbf{rolling learning and evaluation} is as follows. We divide the whole time range of data into non-overlapping intervals, for instance, each interval corresponds to one month (see Figure \ref{fig:eval}). 
We perform the following two steps repeatedly over each interval: (i) we take data within one interval as the testing set and the data in the previous $N$ intervals as the training and validation set,
(ii) each time that the testing data is built from a new interval, the model is retrained and evaluated on the current training and testing data.
Eventually, we obtain the model prediction performance on each testing interval. 
Fig. \ref{fig:eval} provides a toy example. In the top panel, data in interval A and B is sequentially selected as the testing set. Given that $N=2$, the shadow areas show the temporal range of their corresponding training and validation sets. 

For comparison, we also use an \textbf{incremental evaluation} procedure, which amounts to always use all the data preceding to the current testing set for training and validation.
By comparing the prediction performance of models under rolling and incremental procedures, we are able to investigate the robustness of models w.r.t. the look-back time horizon of the training data. Ideally, robust models are able to adapt to the variation in data, thereby leading to comparable performance under different procedures. Previous work using rolling evaluation did not conduct such investigation about the model robustness \cite{jang2018empirical}. 
In particular, the testing set is iteratively selected the same way as the rolling procedure, while the training data is incrementally enlarged by adding the expired test set. 
The bottom panel in Fig. \ref{fig:eval} demonstrates the process of incremental evaluation.




%% file: experiments.tex
\section{Experiments}\label{sec:exp}
In this section, we present a comprehensive evaluation and reasoning behind the results of the approaches. 
\subsection{Dataset} 
We have collected volatility and order book data ranging from September 2015 to April 2017. It consists of $13730$ hourly volatility observations and $701892$ order book snapshots. Each order book snapshot contains several hundreds of ask and bid orders. The maximum number of ask and bid orders in each snapshot are $1021$ and $965$.
For the volatility time series, Augmented Dickey-Fuller (AD-Fuller) test is rejected at $1\%$ significance level and therefore there is no unit root and no need of differencing  \cite{kirchgassner2007introduction}.
Meanwhile, Kwiatkowski-Phillips-Schmidt-Shin (KPSS) test is rejected at $1\%$ significance level as well, which indicates that volatility is non-stationary and could contain local variation \cite{kirchgassner2007introduction, scott2014predicting}.
Seasonal patterns of volatility series are examined via periodogram and the result shows no existence of strong seasonality.

\subsection{Baselines}
The \textbf{first category of statistics baselines} is only trained either on volatility or return time series. 

\textbf{EWMA} represents the exponential weighted moving average approach, which simply predicts volatility by performing moving average over historical ones \cite{hyndman2014forecasting, campbell2007predicting}. 

\textbf{GARCH} refers to generalized autoregressive conditional heteroskedasticity model and is widely used to estimate the volatility of returns and prices in stock and cryptocurrencies \cite{Hansen2005, Katsiampa2017}. 

\textbf{BEGARCH} represents the Beta-t-EGARCH model \cite{harvey2008beta}.
It extends upon GARCH models by letting conditional log-transformed volatility dependent on past values.

\textbf{STR} is the structural time series model \cite{scott2014predicting}. It is formulated in terms of unobserved components via the state space method and used to capture local trend variation in non-stationary time series.

Under this category, we also have plain \textbf{ARIMA} model\cite{kirchgassner2007introduction}. 

The \textbf{second category of machine learning baselines} learns volatility and order book features simultaneously.

\textbf{RF} refers to random forests.
It is an ensemble learning method consisting of several decision trees for classification, regression and other tasks \cite{liaw2002classification}. 
Recently it is used in analyzing financial price data \cite{jiao2017predicting}. 


\textbf{XGT} refers to the extreme gradient boosting \cite{chen2016xgboost}, which is the application of boosting methods to regression trees. It trains a sequence of simple regression trees and then adds together the prediction of individual trees. \cite{alessandretti2018machine} recently uses XGT to model cryptocurrency market. 

\textbf{ENET} represents elastic-net, which is a regularized regression method combining both L1 and L2 penalties of the lasso and ridge methods \cite{liu2010learning}. 

\textbf{GP} stands for the Gaussian process based regression, which has been successfully applied to volatility estimation \cite{brahim2004gaussian, wu2014gaussian}. It is a supervised learning method which provides a Bayesian nonparametric approach to smoothing and interpolation.

\textbf{LSTMs} models volatility history and order book features by two long short-term memory recurrent neural networks (LSTM) and then uses the joint hidden representations to perform volatility forecasting with the ReLU activation \cite{alessandretti2018machine, jang2018empirical}.

As it is still nontrivial to decipher variable importance of multi-variable time series from neural network models \cite{guo2018multi, guo2018interpretable, qin2017dual}, we do not take into account implementing our temporal mixture models based on deep neural networks in the present paper. The aim of the current work is the first step towards fundamentally understanding the data using models with good interpretability. 

\textbf{STRX} is the \textbf{STR} method augmented by adding regression terms on external features, similar to the way of \textbf{ARIMAX}. 

Meanwhile, aforementioned \textbf{ARIMAX} in Sec. \ref{sec:model} falls under this category as well.

For RF, XGT, ENET, and GP methods, input feature vectors are built by concatenating volatility history and order book features \cite{nguyen2015topic}.
They lack the ability to adaptively use data from different sources as well as charactering the time varying interaction.  
Our proposed Gaussian and log-normal temporal mixture are respectively denoted by \textbf{TM-G} and \textbf{TM-LOG} \footnote{The code and data will be public after the paper is accepted.}.

\begin{table*}[!htbp]
  \centering
  \caption{Test errors over each time interval of rolling learning (RMSE)}
  \begin{tabular}{|C{1.5cm}|C{0.9cm}|C{0.9cm}|C{0.9cm}|C{0.9cm}|C{0.9cm}|C{0.9cm}|C{0.9cm}|C{0.9cm}|C{0.9cm}|C{0.9cm}|C{0.9cm}|C{0.9cm}|C{0.9cm}|}
    \hline
    {Interval} & 1 & 2 & 3 & 4 & 5  & 6 & 7 & 8 & 9 & 10 & 11 & 12 \\
    \hline
    EWMA & $0.082^{**}$& $0.136^{*}$& $0.265^{*}$& $0.182^{}$& $0.096^{*}$& 
          $0.027^{*}$& $0.034^{*}$& $0.030^{*}$& $0.056^{*}$& $0.054^{*}$& 
          $0.064^{*}$& $0.096^{*}$\\
    GARCH & $0.136^{**}$& $0.225^{**}$& $0.464^{**}$& $0.286^{**}$& $0.135^{**}$&                                        $0.038^{**}$& $0.046^{**}$& $0.047^{**}$& $0.097^{**}$& $0.104^{**}$& 
           $0.086^{**}$& $0.127^{**}$ \\
    BEGARCH & $0.134^{**}$& $0.223^{**}$& $0.462^{**}$& $0.283^{**}$& $0.133^{**}$&                                        $0.037^{**}$& $0.045^{**}$& $0.045^{**}$& $0.096^{**}$& $0.101^{**}$& 
             $0.083^{**}$& $0.123^{**}$ \\ 
    STR & $0.111^{**}$& $0.207^{**}$& $0.408^{**}$& $0.252^{**}$& $0.105^{**}$& 
         $0.042^{**}$& $0.058^{**}$& $0.031^{**}$& $0.082^{**}$& $0.232^{**}$& 
         $0.065^{*}$& $0.111^{**}$ \\
    ARIMA & $0.106^{**}$ & $0.194^{**}$& $0.409^{**}$& $0.247^{**}$& $0.101^{**}$&                                        $0.117^{**}$& $0.037^{**}$& $0.031^{**}$& $0.084^{**}$& $0.211^{**}$& 
           $0.066^{**}$& $0.096^{**}$ \\
    \hline
    ARIMAX & $0.126^{**}$& $0.282^{**}$& $0.443^{**}$& $0.271^{**}$& $0.134^{**}$&                                         $0.156^{**}$& $0.072^{**}$& $0.041^{**}$& $0.094^{**}$& $0.247^{**}$& 
            $0.076^{**}$& $0.107^{**}$ \\
    STRX & $0.159^{**}$& $0.249^{**}$& $0.414^{**}$& $0.242^{**}$& $0.176^{**}$&                                         $0.125^{**}$& $0.044^{**}$& $0.045^{**}$& $0.104^{**}$& $0.255^{**}$& 
          $0.078^{**}$& $0.138^{**}$ \\
    RF & $0.082^{**}$& $0.151^{**}$& $0.296^{**}$& $0.212^{**}$& $0.096^{**}$& 
        $0.076^{**}$& $0.060^{**}$& $0.029^{*}$& $0.066^{**}$& $0.051^{**}$& 
        $0.061^{*}$& $0.085^{*}$\\
    XGT & $0.076^{*}$& $0.144^{**}$& $0.264^{**}$& $0.194^{**}$& $0.090^{}$&                                        $0.096^{**}$& $0.046^{**}$& $0.031^{**}$& $0.061^{**}$& $0.050^{*}$& 
         $0.061^{**}$& $0.088^{*}$\\
    ENET & $0.080^{**}$& $0.138^{**}$& $0.270^{**}$& $0.184^{**}$& $0.102^{**}$&                                         $0.045^{**}$& $0.035^{**}$& $0.028^{}$& $0.064^{**}$& $0.051^{**}$& 
          $0.059^{}$& $0.084^{}$\\
    GP & $0.082^{**}$& $0.137^{**}$& $0.373^{**}$& $0.196^{*}$& $0.108^{**}$& 
        $0.067^{**}$& $0.042^{**}$& $0.028^{}$& $0.061^{**}$& $0.053^{**}$& 
        $0.063^{**}$& $0.085^{*}$\\
    
    LSTMs & $0.081^{**}$& $0.133^{**}$& $\textbf{0.258}^{}$& $0.190^{}$& $0.101^{**}$&           
         $0.038^{**}$& $0.040^{**}$& $0.030^{*}$& $0.056^{*}$& $0.054^{**}$& 
         $0.063^{**}$& $0.084^{}$\\
    \hline
    TM-LOG& 0.083& 0.186& 0.339& \textbf{0.172}& 0.104& 
            0.033& 0.039& 0.034& 0.098& 0.067& 
            0.081& 0.150 \\
    TM-G& \textbf{0.075}& \textbf{0.118}& {0.259}& 0.189& \textbf{0.089}&                                       \textbf{0.025}& \textbf{0.031}& \textbf{0.027}& \textbf{0.051}& \textbf{0.047}&                                \textbf{0.058}& \textbf{0.083} \\
    \hline
\end{tabular}
\raggedright  Symbols * and ** respectively indicate that 
the error of TM-G in the table is significantly different from the corresponding one at level $5\%$ and $1\%$ (two sample KS test on error distributions).
\label{tab:roll_rmse}
\end{table*}

\subsection{Evaluation set-up}
Smoothing parameter in EWMA is chosen from $\{0.01, 0.1, 0.2, \ldots, 0.9 \}$ via grid search.
In GARCH and BEGARCH, the orders of autoregressive and moving average terms for the variance are both set to one \cite{Hansen2005}. 
In {ARIMA} and {ARIMAX}, the orders of auto-regression and moving-average terms are set via the correlogram and partial autocorrelation. 
For decision tree based approaches including RF and XGT, hyper-parameter tree depth and the number of trees and iterations are chosen from range $[3, 10]$ and $[3, 200]$ via grid search. For XGT, L2 regularization is added by searching within $\{0.0001, 0.001, 0.01, 0.1, 1, 10\}$.
As for ENET, the coefficients for L2 and L1 penalty terms are selected from the set of values $\{0, 0.001, 0.005, 0.01, 0.05, 0.1, 0.3, 0.5, 0.7, 0.9, 1, 2\}$.
In GP, we adopt the radial-basis function (RBF) kernel, white noise, and periodic kernels\cite{brahim2004gaussian}. The hyper-parameters in GP are optimized via maximum likelihood estimation. 

In LSTMs, the size of recurrent and dense layers is chosen over $\{32, 64, 128, 256 \}$ via grid search. Dropout is set to $0.8$. Learning rate is selected from $\{0.0005, 0.001, 0.005, 0.01 \}$, while L2 regularization is added with the coefficient chosen from $\{0.0001, 0.001, 0.01, 0.1, 1.0\}$

In TM-G and TM-LOG, the regularization coefficient is chosen from the set of values $\{0.0001, 0.001, 0.01, 0.1, 1.0\}$. 
The autoregressive order of volatility history is set as in ARIMA, i.e. $l_v = 16$, while the time horizon $l_b$ of order book feature is empirically set to $30$, namely $30$ minutes of order book features before the volatility to predict. 
We also evaluate the effect of $l_b$ in the experiment below. 

We use root mean squared error (RMSE) and mean absolute error (MAE) as evaluation metrics \cite{Hansen2005}, which are defined as follows: $RMSE = \sqrt[]{ \sum_{i}^{} (v_i - \hat{v}_i)^2 /n }$ and $MAE = 1/n \sum_{i} |v_i - \hat{v}_i|$.
Furthermore, we also report the statistical significance of the error distributions of different models by the two-sample Kolmogorov-Smirnov test \cite{murphy2012machine}.

Regarding the evaluation procedure, volatility and order book data are divided into a set of monthly data. Totally, we have $12$ testing months, namely, we retrain models for each testing month and evaluate the performance in the corresponding testing month. In the rolling evaluation procedure, the period of training and validation data prior to a certain testing month is set to $3$ months. 
\begin{table}[htbp]
  \caption{Test errors over each time interval of incremental learning (RMSE)}
  \begin{tabular}{|c|c|c|c|c|c|}
    \hline
    Internal & 7 & 8 & 9 & 10 & 11  \\
    \hline
    EWMA& $0.034^{*}$& $0.030^{*}$& $0.056^{**}$& $0.054^{**}$& $0.064^{**}$ \\ 
    GARCH& $0.046^{**}$& $0.046^{**}$& $0.097^{**}$& $0.105^{**}$& $0.086^{**}$ \\
    BEGARCH& $0.044^{**}$& $0.044^{**}$& $0.095^{**}$& $0.102^{**}$& $0.084^{**}$ \\ 
    STR& $0.041^{**}$& $0.037^{**}$& $0.083^{**}$& $0.188^{**}$& $0.068^{**}$\\
    ARIMA& $0.039^{**}$& $0.031^{*}$& $0.083^{**}$& $0.222^{**}$& $0.067^{**}$\\
    \hline
    ARIMAX& $0.058^{**}$& $0.050^{**}$& $0.154^{**}$& $0.302^{**}$& $0.105^{**}$\\
    STRX& $0.043^{**}$& $0.048^{**}$& $0.138^{**}$& $0.263^{**}$& $0.095^{**}$ \\
    RF& $0.044^{**}$& $0.039^{**}$& $0.059^{**}$& $0.057^{**}$& $0.068^{**}$\\ 
    XGT& $0.035^{*}$& $0.029^{*}$& $0.053^{*}$& $0.050^{*}$& $0.060^{*}$\\
    ENET& $0.038^{**}$& $0.036^{**}$& $0.076^{**}$& $0.058^{**}$& $0.070^{**}$\\
    GP& $0.043^{**}$& $0.041^{**}$& $0.097^{**}$& $0.061^{**}$& $0.069^{**}$\\
    LSTMs & $0.034^{**}$ & $0.031^{*}$ & $0.054^{**}$ & $0.052^{**}$ & $0.060^{*}$\\
    \hline
    TM-LOG& 0.038& 0.033& 0.098& 0.067& 0.082 \\
    TM-G& \textbf{0.030}& \textbf{0.027}& \textbf{0.048}& \textbf{0.047}& \textbf{0.058}\\
    \hline
\end{tabular}
\label{tab:incre_rmse}
\end{table}

\subsection{Prediction performance}
In this part, we first report the prediction performance of one-step ahead prediction and then some example results of multi-step ahead prediction. 
\begin{table}
  \caption{Test error sensitivity to time horizon of order book features (RMSE)}
  \begin{tabular}{|c|c|c|c|c|c|}
    \hline
    {Order book} & 10 & 20 & 30 & 40 & 50  \\
    \hline
    ARIMAX & $0.112^{**}$& $0.125^{**}$& $0.126^{**}$& $0.126^{**}$& $0.135^{**}$ \\
    STRX & $0.138^{**}$& $0.142^{**}$& $0.159^{**}$& $0.157^{**}$& $0.161^{**}$ \\
    RF & $0.081^{**}$& $0.082^{**}$& $0.082^{**}$& $0.083^{**}$& $0.083^{**}$\\
    XGT & $0.077^{*}$& $0.077^{*}$& $0.076^{*}$& $0.077^{}$& $0.076^{}$\\
    ENET & $0.080^{*}$& $0.080^{**}$& $0.080^{**}$& $0.080^{**}$& $0.081^{**}$\\
    GP & $0.095^{**}$& $0.081^{}$& $0.082^{**}$& $0.084^{**}$& $0.085^{}$\\
    LSTMs & $0.081^{**}$& $0.081^{**}$& $0.082^{**}$& $0.081^{**}$& $0.082^{**}$ \\
    \hline
    TM-LOG& 0.082& 0.082& 0.083& 0.083& 0.084\\
    TM-G& \textbf{0.075}& \textbf{0.075}& \textbf{0.075}& \textbf{0.076}& \textbf{0.076} \\
    \hline
\end{tabular}
\label{tab:horizon_rmse}
\end{table}
\begin{table}[htbp]
  \caption{Test errors of 5-step ahead prediction (RMSE)}
  \begin{tabular}{|c|c|c|c|c|c|}
    \hline
    Interval & 7 & 8 & 9 & 10 & 11  \\
    \hline
    EWMA& $0.036^{}$& $0.035^{*}$& $0.077^{**}$& $0.069^{**}$& $0.067^{*}$ \\ 
    GARCH& $0.049^{**}$& $0.046^{**}$& $0.101^{**}$& $0.111^{**}$& $0.089^{**}$ \\
    BEGARCH& $0.048^{**}$& $0.044^{**}$& $0.103^{**}$& $0.107^{**}$& $0.087^{**}$ \\ 
    STR& $0.051^{**}$& $0.037^{**}$& $0.088^{**}$& $0.194^{**}$& $0.069^{**}$\\
    ARIMA& $0.049^{**}$& $0.033^{*}$& $0.089^{**}$& $0.232^{**}$& $0.071^{**}$\\
    \hline
    ARIMAX& $0.062^{**}$& $0.050^{**}$& $0.161^{**}$& $0.303^{**}$& $0.109^{**}$\\
    STRX& $0.049^{**}$& $0.048^{**}$& $0.142^{**}$& $0.268^{**}$& $0.098^{**}$ \\
    RF& $0.040^{**}$& $0.033^{}$& $0.075^{**}$& $0.060^{**}$& $0.067^{}$\\ 
    XGT& $0.058^{*}$& $0.033^{}$& $0.068^{*}$& $0.063^{*}$& $0.069^{*}$\\
    ENET& $0.039^{**}$& $0.033^{}$& $0.079^{**}$& $0.057^{*}$& $0.067^{*}$\\
    GP& $0.048^{**}$& $0.049^{**}$& $0.101^{**}$& $0.061^{**}$& $0.069^{*}$\\
    LSTMs & $0.046^{**}$& $0.033^{}$& $0.075^{**}$& $0.068^{**}$& $0.072^{**}$\\
    \hline
    TM-LOG& 0.039& 0.038& 0.101& 0.069& 0.095 \\
    TM-G& \textbf{0.035}& \textbf{0.032}& \textbf{0.064}& \textbf{0.054}& \textbf{0.065}\\
    \hline
\end{tabular}
\label{tab:step}
\end{table}

\begin{figure*}[!htbp]
\centering
\includegraphics[width=0.8\textwidth]{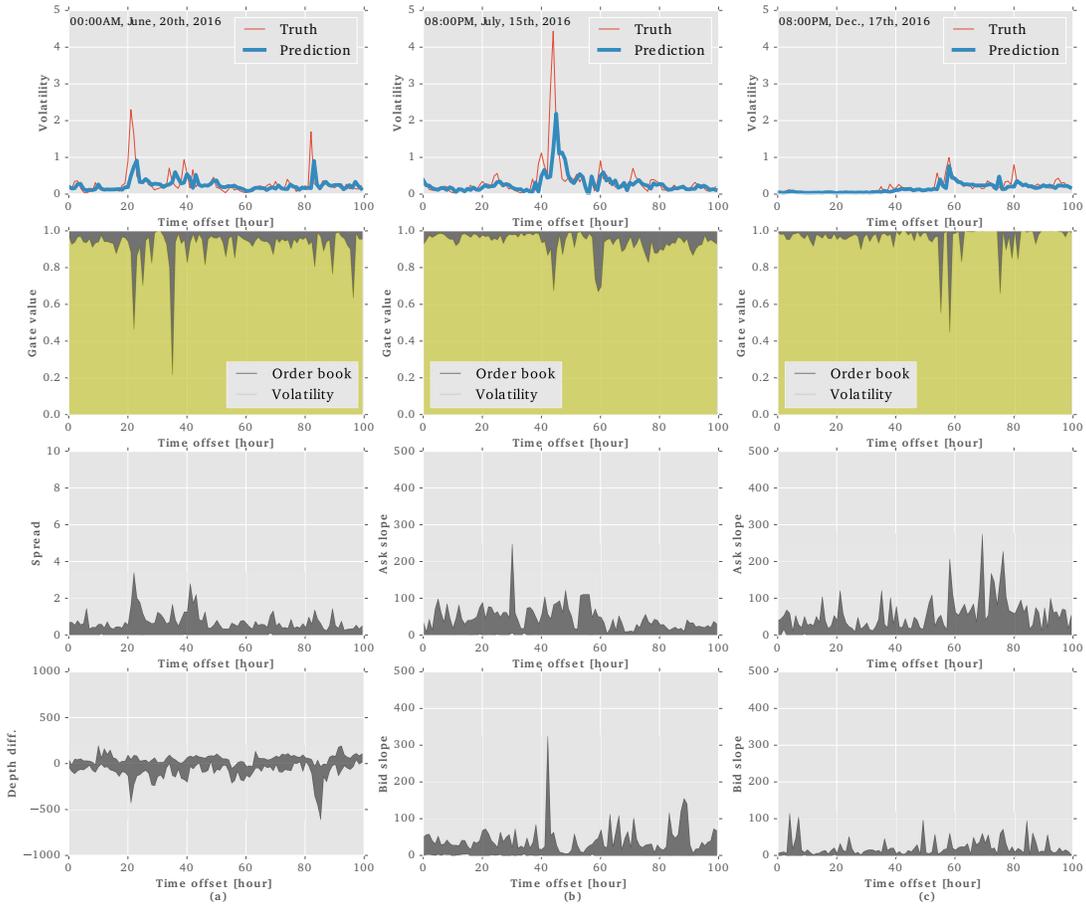}
\caption{Temporal Mixture model visualization. Each column of figures represents the results from the mixture model on a sample testing period, where the time offset is measured in hours w.r.t. to (a)  00:00AM, June, 20th, 2016, (b) 08:00PM, July, 15th, 2016 and (c) 08:00PM, December, 17th, 2016. Top panel: prediction and true values of three sample periods. Second panel: mixture gate value $g_h$ (in yellow color for the weight of volatility history) and $1-g_h$ (in gray color for the weight of order book component model) over time. Bottom two panels: order book feature values over time. (Best viewed in color.)}
\label{fig:model_visual}
\end{figure*}

\textbf{Rolling procedure.}
Tab. \ref{tab:roll_rmse} shows prediction errors over each testing interval obtained by the rolling learning.
The results are reported in three groups of approaches according to the category described in the baseline subsection. 
In general, temporal mixture models, either TM-LOG or TM-G constantly outperform others. Basically, approaches using both volatility and order book features perform better than those using only volatility or return series, however, the simple EWMA can beat many others except for temporal mixture models and LSTMs in some intervals.
Particularly, 
in the middle group of Tab. \ref{tab:roll_rmse},
ARIMAX and STRX using both volatility and order book data fail to outperform their counterparts, i.e. ARIMA and STR in the top group. 
It suggests that simply adding features from order book does not necessarily improve the performance. 
Ensemble and regularized regression perform better, e.g. XGT and ENET perform the best in most of the cases, within this group.  

Gaussian temporal mixture, i.e. TM-G model outperforms other approaches in most of the cases. Specifically, it can achieve $50\%$ less errors at most. 
LSTMs has comparable performance as TM-G in some intervals.
Although the volatility lies only on non-negative range of values, surprisingly TM-LOG is still inferior to TM-G. One possible explanation is that on the short time scales the random variables $v_h \, | \, \{\mathbf{v}_{[h \, , -l_v]} \, , z_h = 0 \}$ and $v_h \, | \, \{\mathbf{X}_{[i(h) \, , -l_b]} \, , z_h = 1 \}$ are better explained with the Gaussian since the variations in short time scale cannot lead to the heavy-tail of the log-normal distribution.

\textbf{Incremental procedure.}
Tab. \ref{tab:incre_rmse} shows the prediction errors obtained by the incremental procedure. 
By comparing the errors in Tab. \ref{tab:incre_rmse} and Tab. \ref{tab:roll_rmse} w.r.t. a certain model and interval, we study the robustness of each model. 
The errors in the front few intervals are comparable to the ones in Tab. \ref{tab:roll_rmse}, because the look-back horizon of training data in the incremental procedure is wider than that in the rolling procedure and the growing size of training data benefits the model training. However, when the horizon further increases, it is interesting to find that such trend changes. 
Due to the page limitation, we list the results from interval $7$ to $11$ in Tab. \ref{tab:incre_rmse}. 

For the models in the top group which do not ingest order book features, the errors of EWMA, GARCH, BEGARCH, and STR are close to the corresponding ones in Tab. \ref{tab:roll_rmse}, though for ARIMA, the errors in interval $10$ and $11$ present little ascending pattern. 
In the middle group, models except LSTMs exhibit decreasing and then increasing error pattern, compared with their counterparts in Tab. \ref{tab:roll_rmse}. 
This suggests that too old data in the training set begins to deteriorate the prediction performance instead.
LSTMs presents robust performance because of the gate and memory mechanism, i.e. errors are comparable over all intervals in Tab. \ref{tab:incre_rmse} and Tab. \ref{tab:roll_rmse}.  
Due to the adaptive weighting mechanism over order book features, TM-G and TM-LOG are robust to increasing amount of training data as well. 
Above observations in Tab. \ref{tab:roll_rmse} and Tab. \ref{tab:incre_rmse} apply to MAE results below as well. 

\textbf{Time horizon of order book feature.}
Tab. \ref{tab:horizon_rmse} demonstrates the effect of time horizon (i.e. $l_b$) of order book features on the prediction performance of models using order book. The results are obtained by evaluating each model on the interval $1$ with increasing size of $l_b$. 
It exhibits that short-term order book features are sufficient for most of the models and furthermore data, e.g. $40$ and $50$ minutes of order book features, leads to no improvement. 
In particular, models like ARIMAX and STRX are prone to overfit by redundant data of long horizon, while our mixture models, LSTMs, XGT, and ENET are relatively insensitive to the horizon.

\begin{table*}[!htbp]
  \centering
  \caption{Test errors over each time interval of rolling learning (MAE)}
  \begin{tabular}{|C{1.5cm}|C{0.9cm}|C{0.9cm}|C{0.9cm}|C{0.9cm}|C{0.9cm}|C{0.9cm}|C{0.9cm}|C{0.9cm}|C{0.9cm}|C{0.9cm}|C{0.9cm}|C{0.9cm}|C{0.9cm}|}
    \hline
    Interval & 1 & 2 & 3 & 4 & 5  & 6 & 7 & 8 & 9 & 10 & 11 & 12 \\
    \hline
    EWMA& $0.047$& $0.063^{}$& $0.158^{*}$& ${0.075}^{*}$& $\textbf{0.050}^{}$& $0.018^{**}$& $0.018^{}$& $0.019^{*}$& $0.026^{*}$& $0.030^{*}$& 
          $0.032^{*}$& $0.040^{*}$\\
    GARCH& $0.099^{**}$& $0.133^{**}$& $0.333^{**}$& $0.147^{**}$& $0.091^{**}$& $0.019^{**}$& $0.029^{**}$& $0.035^{**}$& $0.054^{**}$& $0.075^{**}$& 
           $0.057^{**}$& $0.084^{**}$ \\
    BEGARCH& $0.096^{**}$& $0.131^{**}$& $0.330^{**}$& $0.142^{**}$& $0.087^{**}$& $0.016^{*}$& $0.027^{**}$& $0.033^{**}$& $0.052^{**}$& $0.071^{**}$& 
             $0.053^{**}$& $0.079^{**}$ \\ 
    STR& $0.065^{**}$& $0.105^{**}$& $0.254^{**}$& $0.141^{**}$& $0.051^{}$& $0.028^{**}$& $0.040^{**}$& $0.022^{**}$& $0.036^{**}$& $0.220^{**}$& 
         $0.034^{*}$& $0.065^{**}$ \\
    ARIMA& $0.059^{**}$ & $0.091^{**}$& $0.255^{**}$& $0.123^{**}$& $0.053^{*}$& $0.045^{**}$& $0.020^{**}$& $0.022^{**}$& $0.037^{**}$& $0.204^{**}$& 
           $0.032^{**}$& $0.051^{**}$ \\
    \hline
    ARIMAX& $0.083^{**}$& $0.172^{**}$& $0.295^{**}$& $0.140^{**}$& $0.092^{**}$& $0.050^{**}$& $0.027^{**}$& $0.030^{**}$& $0.045^{**}$& $0.236^{**}$& 
            $0.042^{**}$& $0.062^{**}$ \\
    STRX& $0.103^{**}$& $0.152^{**}$& $0.219^{**}$& $0.163^{**}$& $0.105^{**}$& $0.038^{**}$& $0.047^{**}$& $0.023^{**}$& $0.051^{**}$& $0.175^{**}$& 
          $0.403^{**}$& $0.063^{**}$ \\
    RF& $0.052^{**}$& $0.089^{**}$& $0.174^{**}$& $0.095^{**}$& $0.060^{**}$& $0.071^{**}$& $0.031^{**}$& $0.018^{}$& $0.031^{**}$& $0.031^{**}$& 
        $0.037^{**}$& $0.040^{*}$\\
    XGT& $0.047^{*}$& $0.077^{**}$& $0.159^{**}$& $0.080^{}$& $0.057^{**}$& $0.051^{**}$& $0.031^{**}$& $0.020^{**}$& $0.029^{**}$& $0.031^{**}$& 
         $0.034^{**}$& $0.041^{*}$\\
    ENET& $0.052^{**}$& $0.074^{**}$& $0.166^{**}$& $0.093^{**}$& $0.064^{**}$& $0.036^{**}$& $0.024^{**}$& \textbf{0.017} & $0.028^{**}$& $0.030^{**}$& 
          $0.032^{**}$& $0.035^{*}$\\
    GP& $0.050^{**}$& $0.070^{**}$& $0.216^{**}$& $0.084^{*}$& $0.069^{**}$& 
        $0.055^{**}$& $0.025^{**}$& $0.018^{}$& $0.028^{**}$& $0.034^{**}$& 
        $0.037^{**}$& $0.035^{*}$\\
    LSTMs & $0.048^{*}$& $0.073^{**}$& $0.154^{}$& $0.084^{*}$& $0.059^{**}$& $0.022^{**}$&     $0.030^{**}$& $0.022^{*}$& $0.030^{**}$& $0.032^{**}$& 
         $0.031^{*}$& $0.036^{*}$\\
    \hline
    TM-LOG& 0.053& 0.078& 0.176& \textbf{0.070}& 0.061& 
            0.018& 0.024& 0.041& 0.037& 0.037& 
            0.103& 0.053 \\
    TM-G& \textbf{0.044}& \textbf{0.063}& \textbf{0.153}& {0.079}&                              
          \textbf{0.050}& \textbf{0.014}& \textbf{0.017}& \textbf{0.017}& 
            \textbf{0.024}&  \textbf{0.026}& \textbf{0.029}& \textbf{0.032} \\
    \hline
\end{tabular}
\label{tab:roll_mae}.
\end{table*}
\begin{table}[htbp]
  \caption{Test errors over each time interval of incremental learning (MAE)}
  \begin{tabular}{|c|c|c|c|c|c|c|c|c|c|c|c|c|}
    \hline
    Interval & 7 & 8 & 9 & 10 & 11 \\
    \hline
    EWMA& $0.018^{}$& $0.019^{*}$& $0.026^{*}$& $0.030^{*}$& $0.032^{*}$ \\
    GARCH& $0.028^{**}$& $0.035^{**}$& $0.054^{**}$& $0.075^{**}$& $0.057^{*}$ \\ 
    BEGARCH& $0.025^{**}$& $0.032^{**}$& $0.051^{**}$& $0.071^{**}$& $0.053^{*}$ \\
    STR& $0.022^{*}$& $0.030^{**}$& $0.037^{**}$& $0.179^{**}$& $0.032^{*}$ \\
    ARIMA& $0.020^{*}$& $0.021^{*}$& $0.037^{**}$& $0.215^{**}$& $0.032^{*}$  \\
    \hline
    ARIMAX& $0.037^{**}$& $0.037^{**}$& $0.078^{**}$& $0.291^{**}$& $0.064^{**}$  \\ 
    STRX& $0.048^{**}$& $0.043^{**}$& $0.064^{**}$& $0.195^{**}$& $0.077^{**}$ \\ 
    RF& $0.034^{**}$& $0.031^{**}$& $0.039^{**}$& $0.041^{**}$& $0.048^{**}$ \\ 
    XGT& $0.023^{**}$& $0.021^{*}$& $0.032^{**}$& $0.030^{*}$& $0.032^{*}$ \\ 
    ENET& $0.026^{**}$& $0.025^{**}$& $0.041^{**}$& $0.038^{**}$& $0.052^{**}$ \\ 
    GP& $0.028^{**}$& $0.028^{**}$& $0.050^{**}$& $0.039^{**}$& $0.040^{*}$ \\ 
    LSTMs & $0.025^{**}$& $0.023^{**}$& $0.029^{**}$& $0.033^{**}$& $0.032^{*}$ \\ 
    \hline
    TM-LOG& 0.025& 0.039& 0.038& 0.037& 0.104 \\
    TM-G& \textbf{0.017}& \textbf{0.017}& \textbf{0.024}& \textbf{0.027}& \textbf{0.030} \\
    \hline
\end{tabular}
\label{tab:incre_mae}
\end{table}

\textbf{Multi-step ahead prediction.}
In Tab. \ref{tab:step}, we demonstrate the performance of multi-step-ahead prediction by example results of $5$-step ahead prediction obtained by the rolling procedure. 
Full results will be presented in the future work due to the page limitation.
Basically, in $5$-step ahead prediction all the models present higher errors in comparison to one step ahead prediction in Tab. \ref{tab:roll_rmse}. 
TM-G constantly outperforms baselines. 

\subsection{Model interpretation}
In this part, we provide insights into the data by analyzing
the components of the model. 
Specifically, in Fig. \ref{fig:model_visual} each column of figures corresponds to
a sample period from the testing month. The top panel shows the model prediction and true values of the period. The second panel demonstrates the distribution of gate values in the mixture model corresponding to the top panel. The dark area corresponds to the gate values $1-g_h$ of the component model w.r.t. the order book at each time step. The sum of dark and light areas at each time step is equal to one and therefore the lower the dark area reaches, the higher gate value is assigned to the component model of the order book. 
The time evolution of the inferred gate values $g_h$ and $1-g_h$ in our mixture model explains the dynamical importance and interplay of the order book features for high volatility regimes. Thus, it implies that the order book features can encode the future short-term price fluctuations from the trade orders. 

The bottom two panels exhibit two order book features with high coefficients in the mixture model. It demonstrates the correspondence between feature and gate values over time. Recall that each hourly volatility observation has the associated order book features over a time horizon $l_{b}$ and therefore the value range within $l_{b}$ of each feature at each hour is shown in the figures.
If we look at column (a) the large price fluctuations at the offset of 20 hours from 2016 June, 20th, 00:00 am are mostly driven by the negative market depth feature (panel four) which implies
larger buying demand for the Bitcoins coupled together with the larger spread between bid and ask price (panel three). 
Similarly, the large volatility around the offset of 40 hours from 2016 July, 15th, 08:00 pm in panel (b) is
driven by the large bid slope i.e. buying demand near the currently traded price w.r.t. much smaller ask slope i.e. selling offer near the current traded price. 
In contrast, in panel (c) at the offset of around 60 hours from 2016 December, 17th, 08:00 pm the medium price fluctuations are driven by larger selling offer near the current traded price. These three cases show the interpretability of the dynamical effects in our temporal mixture model to learn the future short-term volatility from the order book. 


%% file: conclusion.tex
\section{Conclusion}\label{sec:conclusion}
In this paper, we study the short-term volatility of Bitcoin market by realized volatility observations and order book snapshots. 
We propose temporal mixture models to capture the dynamical effect of order book features on the volatility evolution for improved prediction and interpretable results. 
We performed comprehensive experiments to compare with numerous statistical and machine learning baselines. 
The proposed temporal mixture models have four favorable properties as: 
(i) it is more accurate in most of the cases than the conventional way of learning data from different sources, which trains models on fused features or learns joint features for prediction. 
(ii) by visualizing the mixture gate values and important features over time, it enables to interpret the effect of order book on the volatility.
(iii) by comparing the prediction performance under rolling and incremental evaluation procedures,
we found out that our temporal mixture model is robust and adaptive w.r.t. time-varying data.  
(iv) it can also serve as a flexible and generic framework for Bitcoin data forecasting and interpretation by adding component models specific to different data sources. 